\documentclass[letterpaper, 10 pt, conference]{ieeeconf}  % Comment this line out if you need a4paper

\IEEEoverridecommandlockouts                              % This command is only needed if
\usepackage{url, amsfonts, amsmath, hyperref, subcaption}
\usepackage[style=numeric,sorting=none,maxbibnames=6]{biblatex}
\usepackage{dblfloatfix}    % To enable figures at the bottom of page
\usepackage{graphicx}

\usepackage{gensymb}
\usepackage{multirow}
\usepackage{MnSymbol}
\usepackage{wasysym}
\usepackage{float}
\usepackage[x11names]{xcolor}
\usepackage{soul}
\usepackage[normalem]{ulem}

\usepackage[labelsep=period,skip=1pt]{caption}
\usepackage{dblfloatfix} 
\usepackage{etoolbox}
\newtoggle{vtwo}
\newtoggle{vtwo_hl}
\newtoggle{vfinal}
\newtoggle{vfinal_hl}

\togglefalse{vtwo}
\togglefalse{vtwo_hl}
\togglefalse{vfinal}
\togglefalse{vfinal_hl}

\usepackage{array, booktabs, makecell} 
\usepackage{siunitx, mhchem}

\title{On the Exploration of LM-Based Soft Modular Robot Design}

\author{Weicheng Ma$^{2,*}$, Luyang Zhao$^{1,*}$, Chun-Yi She$^{1,*}$, Yitao Jiang$^{1}$, Alan Sun$^{1}$, \\ Bo Zhu$^{2}$, Devin Balkcom$^{1}$, Soroush Vosoughi$^{1}$%
\thanks{$^{*}$These authors contributed equally to this work.}%
\thanks{$^{1}$Luyang Zhao, Chun-Yi She, Yitao Jiang, Alan Sun, Devin Balkcom, and Soroush Vosoughi are with the Department of Computer Science, Dartmouth College, Hanover, NH, USA.
        {\tt\small \{luyang.zhao.gr, chun-yi.she.gr, yitao.jiang.gr, Alan.W.Sun.24, devin.balkcom, soroush.vosoughi\}@dartmouth.edu}}%
\thanks{$^{2}$Weicheng Ma and Bo Zhu are with the College of Computing, Georgia Institute of Technology, Atlanta, GA, USA.
        {\tt\small wma76@gatech.edu, bo.zhu@cc.gatech.edu}}%
}

\begin{document}

\maketitle
%\institute{Dartmouth College}

% Designing soft modular robots has traditionally been a labor-intensive task for engineers, requiring extensive trial-and-error experiments and evaluations to determine optimal locomotion strategies in various physical environments. Automating these design tasks poses challenges due to the complexities involved in co-designing the shape module and locomotion control to meet diverse task objectives. We present a novel design system that combines pre-trained language models (PLMs) and differentiable physics simulations to facilitate a fully automated pipeline for synthesizing novel soft modular robot designs. At the heart of our system is a PLM-physics coupling algorithm based on five customizable design metrics to automate the generation and evaluation of new soft modular robot designs adaptable to different physical environments. Specifically, our system processes users' casual natural language inputs, explores feasible modular compositions within the design space by adhering to design constraints, and synthesizes new modularized designs guided by differentiable physics simulation to explore their optimal performance in specific environments. We demonstrate the efficacy of our system by automatically designing soft modular robots capable of uni- and bi-directional locomotion and stair-descending locomotion objectives, and validate their performance with real-world prototypes using engineer-designed gaits.
Recent large language models (LLMs) have demonstrated promising capabilities in modeling real-world knowledge and enhancing knowledge-based generation tasks. In this paper, we further explore the potential of using LLMs to aid in the design of soft modular robots, taking into account both user instructions and physical laws, to reduce the reliance on extensive trial-and-error experiments typically needed to achieve robot designs that meet specific structural or task requirements.
Specifically, we formulate the robot design process as a sequence generation task and find that LLMs are able to capture key requirements expressed in natural language and reflect them in the construction sequences of robots. To simplify, rather than conducting real-world experiments to assess design quality, we utilize a simulation tool to provide feedback to the generative model, allowing for iterative improvements without requiring extensive human annotations.
Furthermore, we introduce five evaluation metrics to assess the quality of robot designs from multiple angles including task completion and adherence to instructions, supporting an automatic evaluation process. Our model performs well in evaluations for designing soft modular robots with uni- and bi-directional locomotion and stair-descending capabilities, highlighting the potential of using natural language and LLMs for robot design. However, we also observe certain limitations that suggest areas for further improvement.

\section{Introduction}

Robots constructed from soft, modular components offer the flexibility to be quickly redesigned or reassembled to adapt to new environments~\cite{ModularSR,softsnap}. However, designing these robots remains a challenge within the robotics community due to the interleaving complexities involved in the combinatorial exploration of modular compositions and the differentiable optimization of control policies. Adapting these robots to various physical environments, characterized by differing terrain, obstacles, and frictional properties, further exacerbates these challenges. Traditional design pipelines rely heavily on extensive trial-and-error experiments to identify viable designs from a vast pool of possibilities, making the design process particularly tedious when tackling robots consisting of many modules in a complex physical environment~\cite{soft_lattice}. Producing a reasonably optimized design remains impractical, particularly for common users, due to the significant engineering and physics expertise required in the design process.

\begin{figure}[htbp]
    \centering
    \includegraphics[width=1\linewidth, trim=0cm 0cm 0cm 3.5cm, clip]{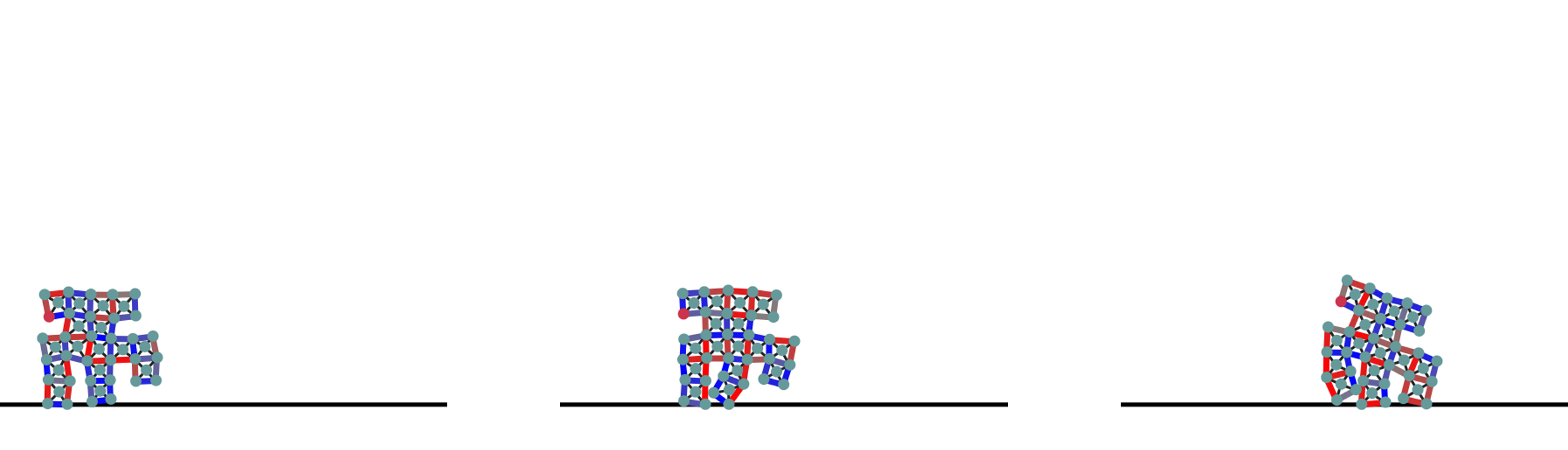}
    \caption{A robot designed by our model with 2.5 feet in contact with the ground}
    \label{fig:oneexample}
    % \vspace{-2em}
\end{figure}

While a few recent works have begun to explore ML-based models for robot design~\cite{Whitman2020ModularRD, RoboGrammar, Hu2022ModularRD}, they often simplify the task by limiting the number of unit robots and predefining the shapes and functionalities of building blocks. 
% Although machine learning (ML)-based robot design models exist [], they often oversimplify the problem by restricting the number of unit robots and predefining the shapes and functionalities of building blocks. 
These predefined elements rely heavily on expert knowledge in robotics, limiting the flexibility and scalability of these models. To our knowledge, none of the modular robot design frameworks have utilized large language models (LLMs), and we are particularly interested in exploring the potential and capabilities of LLMs for this purpose. 
In view of recent advances in the field of natural language processing (NLP) where natural language is used to represent task settings and environments in high-dimensional semantic space~\cite{robonlp-env-1,robonlp-task-1}, this paper explores a new direction
% moves one step further 
by exploring the feasibility of using NLP models to generate robot designs in general, unconstrained scenarios.
This end-to-end setting adopts natural language as the connecting cord throughout the robot designing pipeline, relieving the requirement of expert-level knowledge and introducing high levels of flexibility to the designs.

% Computational design methods present new opportunities to automate these traditionally human-centered design tasks. Machine learning (ML) approaches have been introduced to aid the design process by characterizing the design space with neural networks trained on large datasets and proposing novel designs through their generative capabilities. However, the efficacy of such models is typically limited by the availability of high-quality design examples (i.e., training data) from which the ML models can learn. The creation of this training data still relies heavily on human input and experimental trials, which diminishes the potential benefits of automation.

\begin{figure}[htbp]
    \centering
    \includegraphics[width=1\linewidth]{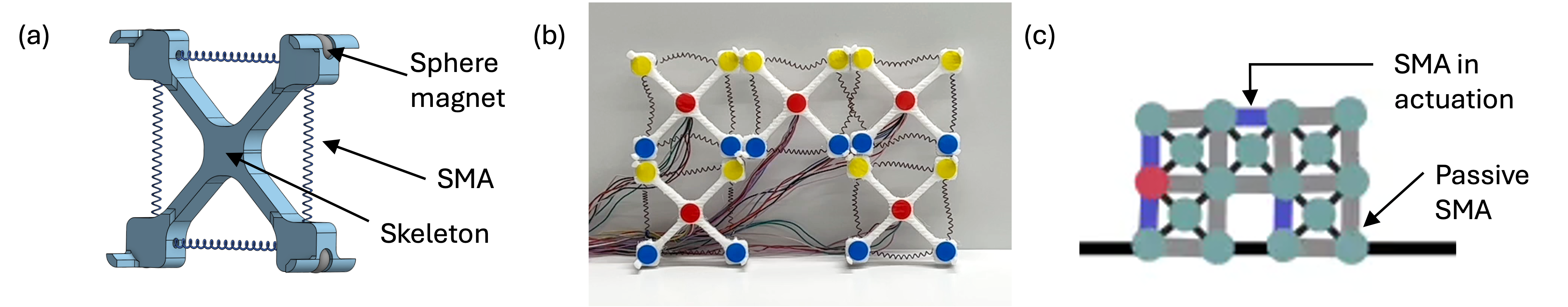}
    \caption{Module design. (a) A single module is composed of a flexible skeleton,
four SMA actuators, and four sphere magnets. (b) An example of real robots built with five modules. (c) A 5-module robot in simulation, where blue lines indicate the springs being actuated, and grey lines indicate the passive SMAs.}
    \label{fig:module_design}
    \vspace{-3.5mm}
\end{figure}

Specifically, we investigate how the expressive flexibility of natural language can represent robot designs, enabling the training of large language models (LLMs) for the robot design task.
% Specifically, we examine the application of the strong expressive power and high freedom of natural language for representing robot designs to enable training large language models (LLMs) for the robot design task.
To overcome the difficulty of supervising the training of ML-based robot generation models, we prepare the training data leveraging automatic data augmentation approaches and use a differentiable physics simulation tool to generate control sequences of the designed robots for the target goals.
Coupled with the simulator, we propose five simulation-based evaluation metrics for automating the evaluation of robot designs in complex physical environments.
These metrics assess the optimality of robot-designing models primarily from the aspects of (1) accommodating the structural requirements in the instructions, (2) producing robots that excel in accomplishing the target task, and (3) designing unseen robots instead of memorizing configurations in the models' training data.
It is critical to note that the choice of the physics simulation tool is arbitrary, and we adopted a differentiable physics simulation tool to reduce the need for human labor in designing the control sequences for the generated robots.
% This paper consists of two main components: (1) We fine-tune a LLM to structure robot designs as natural language prompts and generate combinatorial design choices, supervised by DiffTaichi~\cite{difftaichi-orig}, a differentiable physics simulation tool. (2) We propose five simulation-based evaluation metrics to benefit the future development of soft modular robot designing models. These two components are coupled together and synergistically facilitate the generation and evaluation of novel modular robot co-designs that adhere to physical constraints and meet specific design objectives.
We then conducted different experiments with our LLM-centered robot-designing framework to show the practicability of using natural language to represent and guide the design of soft modular robots.

% \textbf{instruction-following} metric which assesses how well geometric requirements specified in the prompts are addressed, a \textbf{generalizability} metric which calculates the percentage of robot designs not present in the model's training data, a \textbf{success rate} metric indicating the likelihood that the designed robots successfully complete the tasks outlined in the prompts, a \textbf{promise} score which estimates the total distance that the designed robot can travel over a prolonged period, and a \textbf{task optimality} metric which determines the time required for the robots to achieve the task objectives specified in the prompts. These metrics are general-purpose, fitting the evaluation of any robot-designing models with task-oriented configurations mainly for the instruction-following and optimality metrics.  With these metrics in hand, we demonstrate the efficacy of our system by designing a variety of modularized soft robots in \ref{sct:experiments}.

% devin's comment: How large are the robots that the system has designed? Are they large enough to be practical? (No, but maybe they are still interesting.) Are the robots 2d or 3d? What are the limitations? What are some example prompts? Can we see some figures of the generated designs doing things? These things must all be in the introduction. Otherwise it's just an ad.
Our proposed model is trained using 3,000 randomly selected robot configurations and prompted to generate robot designs given specific environments, target goals, and suggestions.
We then match the robots designed by our model with real robots to validate the generations and learn better empirical designs from high-quality designs.
To enable such matching, we designed a 2D real robot that is both soft and modular, as depicted in Figure~\ref{fig:module_design}(a), to form the generated robot design.
Figure~\ref{fig:module_design}(b)(c) displays examples of the real robot and its simulated counterpart.
From our primary experiments, we discovered multiple generations that differ from traditional human-designed robots but perform well toward various objectives in simulations.
For instance, our model designed a robot with two long legs and a short limb (illustrated in Figure~\ref{fig:ownmodel:subfig1}) to achieve forward locomotion, the design of which is not intuitive.
However, this design locomotes efficiently on the flat plane, benefiting from the short limb alternatively serving as a leg or an arm depending on the current deformation state (Figure~\ref{fig:oneexample}).
Such new design is a departure from the traditional three-legged robot designs and motivates domain experts to come up with less straightforward but effective designs.

\begin{figure*}[htbp]
    \centering
    \includegraphics[width=.9\linewidth]{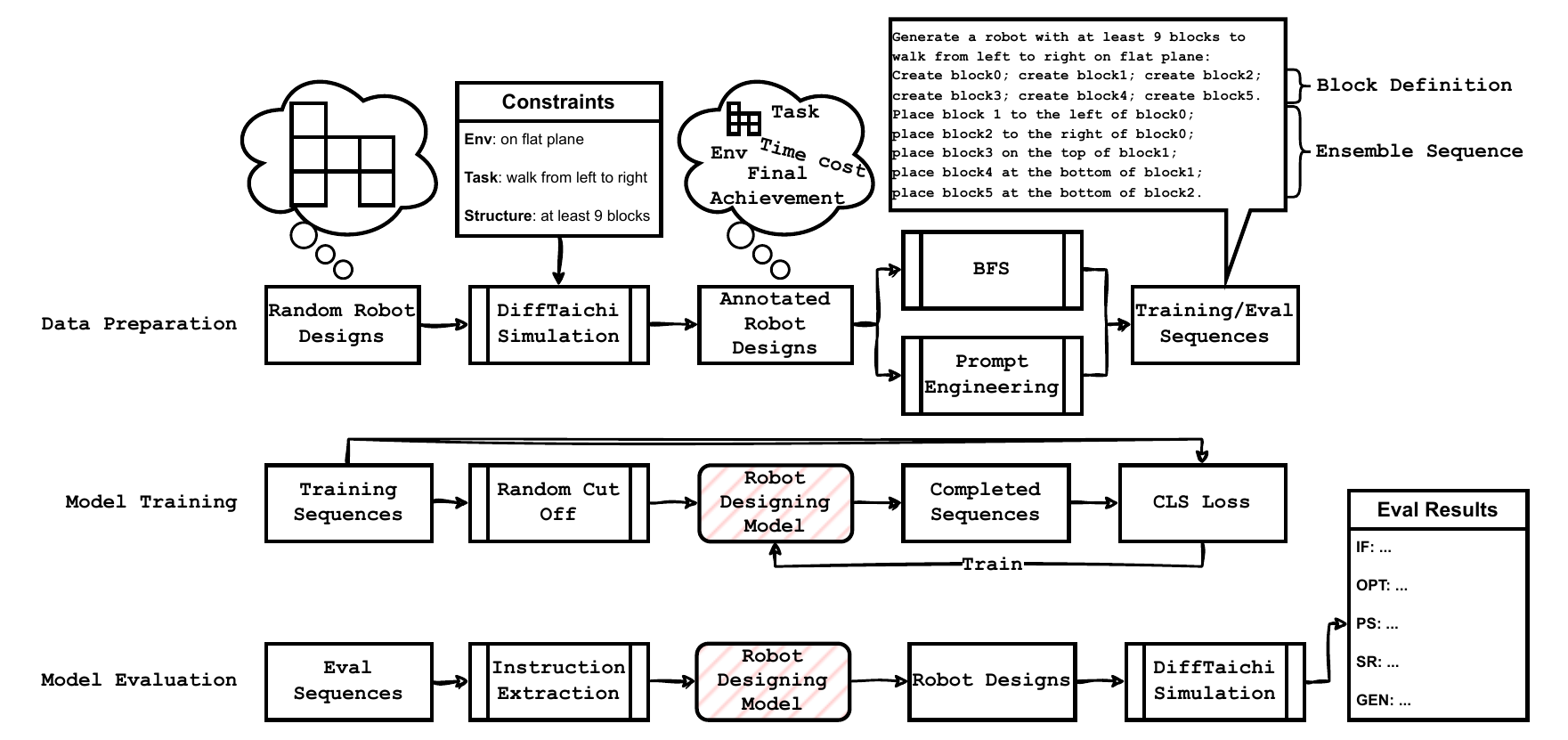}
    \caption{An overview of the data preparation and model training and evaluation pipelines.}
    \label{fig:methods}
    \vspace{-6mm}
\end{figure*}
% \begin{figure*}[htbp]
%     \centering
%     \vspace{-2em}
%     \includegraphics[width=1\linewidth]{figures/robotics_design_construction.png}
%     \caption{We generate design sequences for soft modular robots using breadth-first search (BFS). Each robot corresponds to multiple design sequences. }
%     \label{fig:methods:robot-representation}
%     \vspace{-4mm}
% \end{figure*}

\section{Related Work}

\noindent \textbf{Modular Self-reconfigurable Robots (MSRRs):}
MSRRs present a significant departure from traditional construction methods. These robots are made up of multiple modules, each possessing the ability to move and mechanically connect with one another\cite{M-blocks, roombot_epfl}. This enables them to restructure and adapt to a variety of tasks~\cite{yim2007modular}. Historically, these modules have been rigid. However, the introduction of compliance and flexibility to these modules has expanded the capabilities of soft modular robots. This evolution offers new modes of actuation, mobility, and even safer interactions with humans~\cite{ModularSR,soft_lattice, zappetti2017bio, untethered_isoperimetric}.

\noindent \textbf{Soft Modular Robots:} These robots, with their increased flexibility, can perform a variety of functions by deforming into different shapes from various initial configurations~\cite{ModularSR,mintchev2012underwater, kurumaya2018modular, ze2022soft, li2022scaling, robertson2017new, cagdas2012bend, soBL, kirstin2021swarm, magnet_cube, shrink_cube}. For instance, Zhao et al. \cite{10146508} developed identical self-reconfigurable blocks called Starblocks, which can achieve diverse forms of locomotion by assembling into configurations such as a self-assembled wheel or a quadruped. Additionally, these blocks can be configured into a robotic arm with a gripper for prehensile manipulation, a lattice for non-prehensile manipulation, or even a tent-like structure for formation tasks. Configurations have spanned from simple chain-like formations to sophisticated designs, all aiming to support activities like locomotion, manipulation, and transformation.

% , including diverse forms of locomotion, such as locomoting like a  four legges ~\cite{cagdas2012bend, mintchev2012underwater, kirstin2021swarm,10146508,10146508}, manipulation techniques~\cite{kurumaya2018modular,mckenzie2019linbots,10146508,10146508}, and reconfiguration abilities~\cite{li2022scaling,10146508,10146508}. 

% However, a significant limitation persists: design of the assembly configurations for various tasks still rely heavily on manual assembly, either fully or during the initial stages~\cite{}. 
% % Importantly, none of these robots have been tailored for architectural assembly, thus lacking in 3D lattice stacking features.

\noindent \textbf{Robot Design through Manual Assembly and Iterative Testing:}
In both rigid and soft self-reconfigurable modular robots, the majority of contemporary research still necessitates the manual design of distinct assembled configurations to fulfill various tasks. Though these methodologies prove effective, they often involve significant labor and multiple iterations of trial and error to pinpoint the optimal configuration~\cite{lee2017soft,10146508,li2022scaling}. 
% For instance, Zhao et al.\cite{10146508} designed different assembled configurations like tent, dog, wheel, and robot arm with different robot layout. 
Automating the design of these configurations for various environments and tasks is a crucial area for exploration.

\noindent\textbf{Robot Design Using Learning-Based Techniques}: 
There are several studies that using learning-based methods to design rigid modular robots with different types of components.
Whitman et al. \cite{Whitman2020ModularRD} apply deep reinforcement learning to design modular serial manipulators for specific tasks. Specifically, the study uses 11 types of modular components, including three base mount orientations, one actuated joint, six different links/brackets, and one end-effector. There are constraints between those modular components, such as the maximum number of modules allowed in a configuration, limited to 16, and the need to ensure physical and functional compatibility between the components, which dictates how they can be arranged to achieve the desired task performance.
RoboGrammar, done by Zhao et al., introduce an automated framework that generates optimized robot structures by leveraging a recursive graph grammar and various robot components, including body segments, limbs, and joints, to navigate diverse terrains effectively~\cite{RoboGrammar}. 
Hu et al.~\cite{Hu2022ModularRD} investigate robot design using a generative adversarial network (GAN) called RoboGAN, which works with modular robots composed of different types of modules: body, legs, wheels. RoboGAN learns a mapping from tasks (e.g., terrain types) to a distribution of modular robot designs. Unlike traditional methods that typically produce a single optimal design, RoboGAN can generate multiple distinct designs that are all viable for the given task.

% However, all related works to date have focused on exploring robot designs composed of rigid, function-specific modules, which offer relatively lower degrees of freedom compared to soft modular robots. To our knowledge, there has been no research on the design of soft modular robots. Our work addresses this gap by focusing on robots made of identical soft modules, which can be connected in any direction, free from arrangement limitations. This flexibility allows them to be assembled into different configurations to accomplish various tasks. The design space for these robots is vastly larger and not predefined, requiring a generative model to handle the complexity and scale of the design possibilities. Additionally, the use of Large Language Models (LLMs) in robot design has not been explored. Our research aims to investigate the potential of LLMs for designing soft modular robots, offering a new direction for further study.

However, all related works focus on designs composed of rigid, function-specific modules, which offer relatively lower flexibility compared to soft modular robots. To our knowledge, there is no research on designing soft modular robots. Our work addresses this gap by focusing on identical soft modules that can connect freely in any direction, supporting diverse configurations for various tasks. This design space is vastly larger and not predefined, requiring a generative model to manage its complexity. Additionally, the use of Large Language Models (LLMs) in robot design remains unexplored. Our research aims to investigate LLMs for designing soft modular robots, offering a new direction for further study.

% We are also investigating the potential of using Large Language Models (LLMs) for designing soft modular robots. This represents a first step in exploring the capability of LLMs for robot design.

% The work in \cite{varshavskaya2008automated} delved into merging reinforcement learning with the design of distributed controllers for modular reconfigurable robots, which are envisioned as rigid blocks that are immobilized when interconnected and can only move one by one blocks.

\noindent \textbf{Differential-based Simulation Tools:}
The use of differential-based simulation tools, such as diffTaichi, has become commonplace in robot gait generation. These tools offer a virtual environment where various robot configurations can be tested for their efficacy, eliminating the need for resource-intensive physical prototypes \cite{difftaichi-orig}. Simulation tools have been particularly instrumental in advancing soft modular robot research by offering insights into potential design adaptations before actual assembly.

\section{Methods} 
This paper explores the use of LLMs for designing soft modular robots, employing a natural language generation input and output style.
% As depicted in Figure \ref{fig:methods}, our system comprises three main phases: training data preparation, model training, and model evaluation.
The following subsections provide details on the configuration and training/evaluation processes of our proposed model.

\subsection{Robot Designing Model}
We adopt a generative LLM to design robots with pre-specified environments, task objectives, and structural requirements.
Input and output formats of the model are specified in Sections \ref{sct:method:env-targets}-\ref{sct:method:robot}, respectively.

\subsubsection{\normalfont Environment and Task Representation} \label{sct:method:env-targets}
Leveraging the LLMs' proficiency in understanding natural language, we craft the prompts in natural language, instructing the model to \textit{``Design a soft modular robot for [task-objective] for [distance] in [environment] using [structural-constraints].''} Here, {\em [task objective]} and {\em [environment]} are essential parameters that link the model to the simulation tool, enabling the model to learn to generate legal and high-quality robot designs.
{\em [Distance]} and {\em [structural constraints]} are optional parameters introducing variability into the prompts and aid in minimizing the risk of overfitting. Below are two example prompts that share the same task objective and environment but vary in their specifications:
\\
\textit{``Design a soft modular robot for walking from left to right on a flat plane using at most 9 blocks.''}\\
\textit{``Design a soft modular robot for walking from left to right for at least 6 body\_length on a flat plane.''} 

To demonstrate the workflow, we design three specific task objectives: unidirectional and back-and-forth locomotion on a horizontal plane, and stair-descending locomotion. 

\subsubsection{\normalfont Design Representation} \label{sct:method:robot}
The design representation consists of two parts, i.e., a block-definition section and an ensemble sequence section.
Specifically, the block definition section specifies the properties and constraints of the individual blocks that will be used in the robot's construction. The robot ensemble section then provides a step-by-step description of how these blocks are assembled to form the complete robot, detailing the spatial and functional relationships between the blocks to achieve the desired design.
Since each robot design could be assembled in arbitrary orders, it could be mapped to multiple language representations, enabling easy data augmentation via breadth-first search (BFS).
An example design of a robot and one possible language representation of it is illustrated in Figure \ref{fig:methods}.

\subsection{Evaluation Metrics and Settings} \label{sct:method:eval}
We define five reference-less metrics to evaluate soft modular robot designing models from the instruction-following, optimality, and novelty aspects. \textbf{(1) The instruction-following metric (IF)} assesses how closely a model's generated robot designs adhere to the structural requirements specified in the prompts. \textbf{(2) The promise score (PS)} estimates the total locomotion distance that each robot designed by the model can achieve within a long duration. \textbf{(3) The task optimality metric (OPT)} gauges the efficiency of the robot designs by measuring the time needed to complete the tasks specified in the prompts. \textbf{(4) The generalizability metric (GEN)} calculates the percentage of robot designs not previously seen in the model's training data, highlighting the model’s ability to improvise instead of memorize. \textbf{(5) The success rate (SR)} measures how frequently the model can successfully generate a legal robot configuration description.

% In our evaluation, the model receives the same set of five prompts given each specific task objective and repeatedly generates 10 robot designs for each prompt to mitigate variability. The results are then averaged across all the queries for each metric. 
Note that in this study, the capabilities of our framework are constrained by the simulation tool, DiffTaichi, limiting our testing to relatively simple locomotion tasks. Consequently, we tailored the \ul{OPT} metric to these tasks and performed calculations based on the time it takes for a robot to complete a given task. As we expand our framework to incorporate more advanced simulators and diverse task types, the design of the OPT metric will also be adapted to align more closely with the specific objectives of these new tasks.

%Since SR, PS, and OPT are simulation-based metrics, a high-quality simulator is a fundamental part of the metric. Although the simulator is not expected to mimic real-world conditions perfectly, it has to closely align with the outcomes of real-robot experiments. This alignment is substantiated in Section \ref{sct:sim-to-real}, where simulation results using DiffTaichi demonstrate high consistency with real-robot experimental data, thereby providing a reliable basis for evaluating our robot design models.

\subsection{Training Data Preparation} \label{sct:method:data-prep}
The training of our model does not rely on human-annotated data but instead utilizes synthesized data generated through DiffTaichi. Specifically, we engineer task objectives and environments, randomize parameters such as the distance between the starting point and the destination or stairs, and simulate movements across various robot configurations. After these simulations, the generated data points are formatted as tuples of (task objective, environment, maximum distance, robot configuration, and time cost) to form the training dataset.
The robot configurations are natural language descriptions of robots, expressed using the format exemplified in Figure \ref{fig:methods} and diversified using BFS.
% As mentioned above, BFS is used to enhance the diversity of the robot configuration description in natural language description. 
%The robot configurations are natural language descriptions of robots, expressed using the format exemplified in Figure \ref{fig:methods:robot-representation}.
%To enhance the diversity of these descriptions, we employ BFS to create multiple natural language descriptors of each robot, as detailed in Figure \ref{fig:methods:robot-representation}. 
In our principal experiments, our training data involves 3,000 randomly sampled robot configurations within $5\times 5$ grids. 

Leveraging the training data, we devise two categories of natural language prompts for training our model, targeting causal language modeling (CLM) \cite{clm-1} and binary decision-making objectives. 

\subsubsection{\normalfont CLM Objective}
For the CLM objective, data point entries are incorporated into a predefined sentence template, i.e., \textit{``Design a soft modular robot to achieve [task-objective] over a distance of [min/max distance] within [environment] using [min/max number-of-blocks] blocks.''}
To enhance the diversity of the training prompts and minimize the risk of overfitting, the inclusion of the \textit{distance} and \textit{number-of-blocks} parameters is randomized. 
% When included, the  \textit{[distance]} parameter is specified as \textit{``at least [random reachable distance],''} and the \textit{[number-of-blocks]} parameter is expressed either as a minimum \textit{(``at least'')} or a maximum \textit{(``at most'')}, based on randomly selected numbers appropriate for the actual number of blocks in each robot configuration.

\subsubsection{\normalfont Decision-making Objective}
For the binary decision-making objective, we randomly pair data points designed for the same environment and task objective (designated as [data1] and [data2]) to create comparative prompts structured as follows: \textit{``For achieving [task-objective] over a distance of [min/max distance] within [environment] using [min/max number-of-blocks] blocks, which design is better? (a) [robot configuration of data1]. (b) [robot configuration of data2]. [robot configuration of winner].''} Here, the ``winner'' is determined as the configuration with the lower time cost. 
% Our model is trained to complete these texts, with the initial part of each prompt fixed as input up to a point indicating the superior design. 
Similar to the CLM objective, the inclusion of the \textit{distance} and \textit{number-of-blocks} parameters is made optional.

\section{Robot Design and Calibration of Simulation}

This section outlines the design of our two-dimensional (2D) robotic model and the subsequent calibration of its simulation counterparts. The calibration process is driven by the utilization of empirical data from real-world robot operations, ensuring that the simulations accurately reflect the behaviors observed in the actual robots.
% System identification plays an important role in bridging the gap between the physical behaviors of real robots and their digital counterparts in simulation environments. 

\subsection{Robot design}

\begin{figure}[h]
    \centering
    \includegraphics[width=1\linewidth]{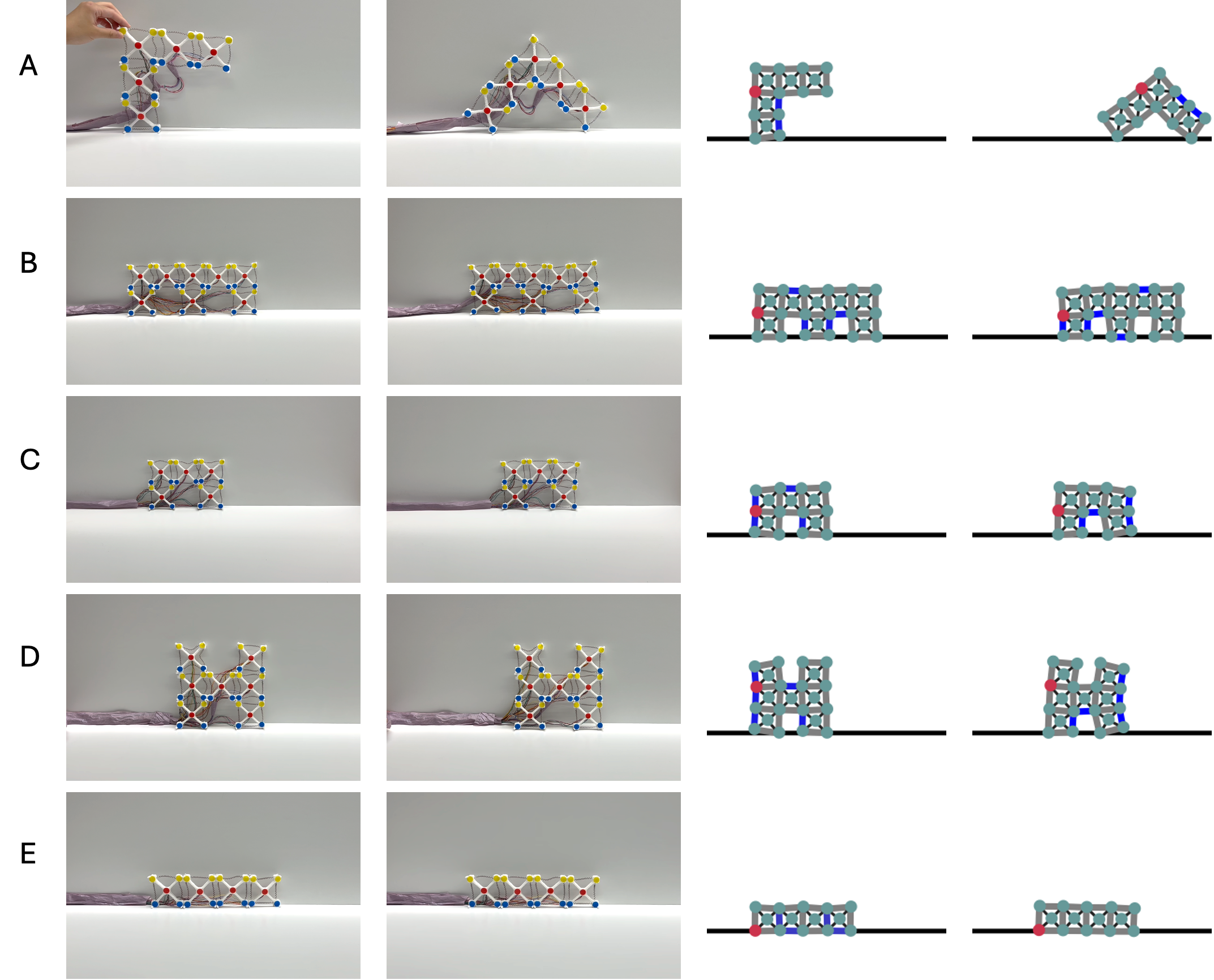}
\caption{Examples of locomotion for five different robot configurations in both the physical world and simulation. For each row, from left to right: initial configuration in the real world, final configuration after four rounds of gait, initial configuration in the simulation, and final configuration in the simulation.}
    \label{fig:real-sim}
    \vspace{-1em}
\end{figure}

An individual module (Figure~\ref{fig:module_design}(a)) consists of a skeleton, four shape memory alloy (SMA) coils (coil diameter: 3.45 mm; wire diameter: 0.51 mm, Dynalloy), and spherical magnets (diameter: 3/8"). The skeleton has an 'x' shape that matches the shape used in the simulator, with four equal-length bars connected at the center. It is printed from flexible TPU material with an infill density of 100\% to provide a stronger adversarial bending force, helping restore the module's original shape after SMA contraction. The SMAs are mounted between the adjacent bars.

Each SMA spring is crimped with fishing wire and electric wire in a ferrule and attached to grooves in the skeleton. The detwinned Martensite rest length of each SMA is longer than the distance between the grooves, allowing extension when the skeleton deforms due to SMA actuation. The edge length of each module is about 7.2 cm. The Austenitic (actuated) rest length of the actuators is 1.2 cm, while the detwinned Martensite rest lengths are set to 3.7 cm upon installation. These lengths can change with actuation cycles due to variations in antagonistic forces from other SMA actuators and the elastic energy stored in the deformed soft skeleton.
When the SMAs are actuated, they deform the skeleton. As they cool, the restoring force of the skeleton stretches the SMAs back to their detwinned Martensite rest length. We control the SMAs in an open-loop manner by actuating them for a specific duration under 5V. Physical experiments were conducted for each application to determine the appropriate actuation duration. Each SMA coil needs 15-20 seconds to cool down and recover.

Spherical magnets are attached to the ends of each limb of modules to allow them to connect with each other. The polarity of these magnets was set according to the method outlined in \cite{Zwier2017MagneticSC}, where four magnets form a saddle configuration. A saddle of magnets is a polygon with alternating polarities, where magnets point alternately inward and outward. By defining the top and bottom directions of each robot and aligning the magnetic spheres accordingly, any module can be connected from any direction.

\subsection{Calibration of Simulation} \label{sct:sim-to-real}

To ensure the simulation reflects the behaviors observed in real robots, we manually designed five distinct configurations (shown in Figure~\ref{fig:real-sim}: a linear chain, a 5-robot two-legged configuration, a 7-robot two-legged configuration, an 8-module three-legged configuration, and a 5-module L-shape configuration. 
These configurations were chosen to explore the effects of different aspect ratios, the number of legs, and symmetry on locomotion. The L-shape configuration, in particular, was selected because our initial simulations with DiffTaichi and LLM learning showed a preference for configurations that could initially topple due to an unbalanced center of mass. We aimed to investigate how such configurations perform in the real world.

\begin{figure*}[htbp]
    \centering
    \begin{subfigure}[b]{0.19\textwidth}
        \centering
        \includegraphics[width=\textwidth]{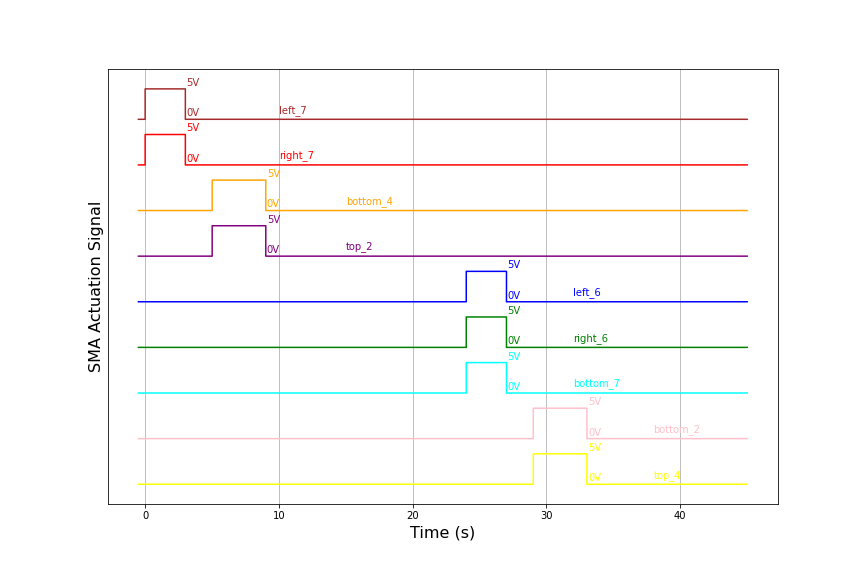}
        \caption{5-module 2-leg}
        \label{fig:controlsequence_5}
    \end{subfigure}
    \hfill
    \begin{subfigure}[b]{0.19\textwidth}
        \centering
        \includegraphics[width=\textwidth]{figures/actuation_signals-8-robots.png}
        \caption{5-module 'L'-shape}
        \label{fig:controlsequence_L}
    \end{subfigure}
    \hfill
    \begin{subfigure}[b]{0.19\textwidth}
        \centering
        \includegraphics[width=\textwidth]{figures/actuation_signals-8-robots.png}
        \caption{4-module 1-chain}
        \label{fig:controlsequence_4}
    \end{subfigure}
    \hfill
    \begin{subfigure}[b]{0.19\textwidth}
        \centering
        \includegraphics[width=\textwidth]{figures/actuation_signals-8-robots.png}
        \caption{7-module 2-leg}
        \label{fig:controlsequence_7}
    \end{subfigure}
    \hfill
    \begin{subfigure}[b]{0.19\textwidth}
        \centering
        \includegraphics[width=\textwidth]{figures/actuation_signals-8-robots.png}
        \caption{8-module 3-leg}
        \label{fig:controlsequence_8}
    \end{subfigure}

    \caption{Control sequences for various robot configurations: (a) 5-module 2-leg, (b) 5-module 'L'-shape, (c) 4-module 1-chain, (d) 7-module 2-leg, and (e) 8-module 3-leg.}
    \label{fig:control_sequences_combined}
    \vspace{-0.8em}
\end{figure*}

For each configuration, we manually designed the gaits for locomotion (i.e., the control sequence of SMA actuations, detailed control signal can be found in Fig.~\ref{fig:control_sequences_combined}).  We applied the same gaits and actuate the same round of gaits to both physical robots and simulations and used the distance traveled relative to the body length of the module as the calibration parameter.
Our results show that with parameter fine-tuning in the simulation, the real-world robot and simulation exhibited the same ranking among these five configurations as shown in Fig.~\ref{fig:real-sim}. The ranking, in both the real robot and simulation, was as follows: 5-module L-shape $>$ 8-module three-legged configuration $\approx$ 5-robot two-legged configuration $>$ 7-robot two-legged configuration $>$ linear chain. 
We also noticed that while the L-shape ranked first in both cases, it moved faster in simulation due to inertia and a bouncier skeleton, which doesn’t match the real robot and is difficult to adjust in DiffTaichi. Thus, we excluded configurations prone to tipping from the training cases.
% which doesn't fit with the real robot and cannot be easily modified in DiffTaichi. Therefore, in the training cases, we removed the instances where the configuration could fall at the beginning due to unbalanced weight.

Moreover, the real robots are actuated using SMAs, which require a long cooling time to return to their original shape. However, in the simulation using DiffTaichi, the actuators return to their original shape instantaneously. To align the simulation with the real robot's behavior, we adjusted the control sequence in the simulation by removing the long waiting time necessary for the real robot. For the real robot, we maintained a consistent cooling period after each control action. This adjustment was applied uniformly across all configurations in the simulation.

\begin{table}[t]
\centering
\small
\setlength{\tabcolsep}{3.2pt}
\begin{tabular}{ccccc}
\hline
\multicolumn{1}{c|}{\textbf{IF}} & \multicolumn{1}{c|}{\textbf{PS}} & \multicolumn{1}{c|}{\textbf{OPT}} & \multicolumn{1}{c|}{\textbf{GEN}} & \textbf{SR} \\ \hline
\multicolumn{5}{c}{\textbf{Uni-Directional Locomotion}}                                                                                                             \\ \hline
\multicolumn{1}{c|}{70.00\%}     & \multicolumn{1}{c|}{13.78}        & \multicolumn{1}{c|}{5.45}         & \multicolumn{1}{c|}{81.00\%}     & 98.00\%      \\ \hline
\multicolumn{5}{c}{\textbf{Back-and-Forth Locomotion}}                                                                                                              \\ \hline
\multicolumn{1}{c|}{61.00\%}     & \multicolumn{1}{c|}{11.99}        & \multicolumn{1}{c|}{2.19}         & \multicolumn{1}{c|}{74.00\%}     & 89.00\%      \\ \hline
\multicolumn{5}{c}{\textbf{Downstairs Locomotion}}                                                                                                                  \\ \hline
\multicolumn{1}{c|}{61.00\%}     & \multicolumn{1}{c|}{16.73}        & \multicolumn{1}{c|}{4.11}         & \multicolumn{1}{c|}{71.00\%}     & 97.00\%      \\ \hline
\end{tabular}
\caption{Evaluation results of our model in three locomotion tasks.}
\label{tbl:experiment:result}
\vspace{-6mm}
\end{table}

\begin{figure}[t]
    \centering
    \includegraphics[width=.4\linewidth]{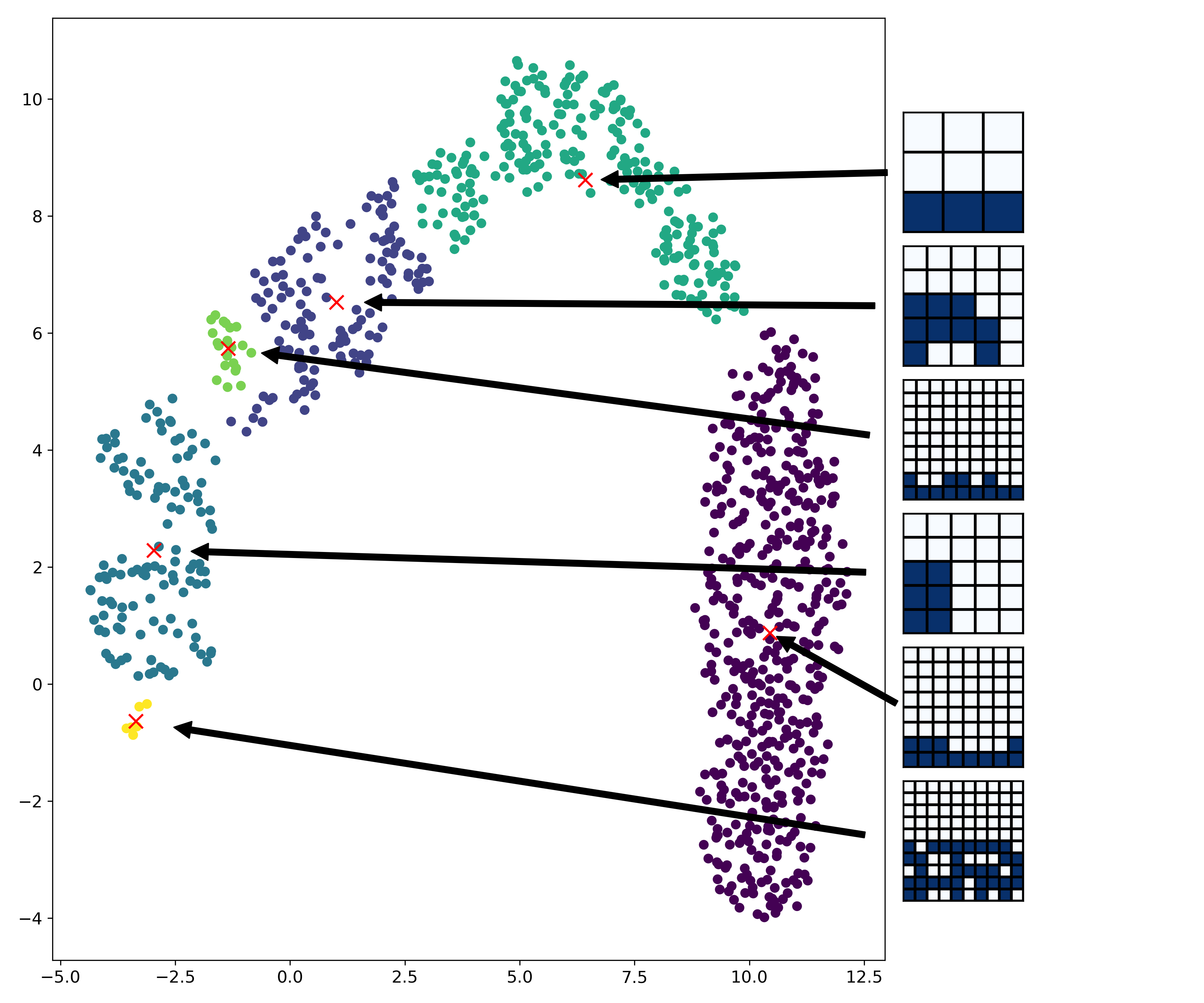}
    \caption{Umap visualization of diverse robot designs encoded by our proposed model. Representative robot designs from the center of each cluster are also displayed.}
    \label{fig:experiments:umap}
    \vspace{-6mm}
\end{figure}

\section{Experiments} \label{sct:experiments}
We evaluate the performance of our proposed model using the five quantitative metrics outlined in Section \ref{sct:method:eval}, complemented by qualitative analyses of representative generations. We use five prompts for each task objective and generate 10 robot designs per prompt for quantitative evaluation.
% [Matt TODO]: Specify parameters in the metric
The model employs a GPT-NEO \cite{gpt-neo} backbone and is trained using synthetic data across all three task objectives introduced in Section \ref{sct:method:env-targets}. We aggregate the scores for each task objective and present them in Table \ref{tbl:experiment:result}.

% IF + SR
In response to all prompts, our proposed model consistently generates legal robot designs, ensuring that all modules are correctly interconnected without any ``floating'' or detached ones indicated by the relatively high success rate (SR) (over 89\%) score. The instruction following (IF) score of over 60\% shows the majority of robots designed by our model satisfy the structural and task-oriented requirements in the prompts.

% The evaluation results underscore the model's adherence to the specified prompts, indicated by the high IF and SR scores.
% where the generated robot designs precisely satisfy the requirements in the prompts in over 90\% cases. Furthermore, the success rate (SR) scores always surpass 50\% for these objectives, indicating that the majority of the robot configurations generated by our model meet the design requirements. 
% This performance not only confirms the effectiveness of our metrics in identifying designs that fall short but also highlights the precision of our model in crafting task-specific robotic structures that are both functional and compliant with design criteria. 
% Figure \ref{fig:experiments:results} showcases both successful and suboptimal examples from the evaluations.
% PS + OPT
The task optimality (OPT) and promise score (PS) metrics further show that the designed robots can complete the tasks within reasonable timeframes and have the capacity to travel longer distances than required within simulation time(on average 14.16 block lengths).
% These findings affirm the model’s ability to produce relatively high-quality robot designs as instructed. 
% GEN
For the generalizability (GEN) metric, which evaluates the creativity of robot-designing models, our model scores between 71\% and 81\%. This range indicates that the model generates robots distinct from the training data most of the time.
Such results suggest our model's ability to infer the geometric and structural characteristics of superior robots rather than merely memorizing good robot designs in the training data.
% [Matt&Luyang TODO]: additional testing with the 3-objective model, give it a yet unseen or unverifiable objective (like going upstairs and then downstairs) and explain the reasonability & generalizability of the model via simulation (if possible) and manual analysis (e.g., what makes the difference between the model's designs in seen vs. unseen objectives; whether the new designs make sense, etc.)

To verify that the geometric and structural information of robots is captured and utilized by our model when producing robot designs, we use UMAP \cite{umap-orig} to visualize the encoding of robot designs in the latent encoding space of our model.
Our guiding hypothesis is that the similarity in the encodings of two robots indicates the model's perceived similarity between them.
To preclude information leakage stemming from the binary choices training objective, we randomly select 1,000 robots not included in our model’s training dataset and generate UMAP visualizations for these sampled robots.
The UMAP visualization is displayed in Figure \ref{fig:experiments:umap}, with one representative robot design shown on the side of each cluster to reveal the clustering evidence.

The visualization indicates that our model effectively encodes the geometric characteristics of robot designs, e.g., the model distinctly clusters robots with varying aspect ratios within its latent space, suggesting the potential of our model to generate robot designs within complicated environments such as crawling through a small hole. Additionally, our model exhibits understandings of critical design elements such as the presence of legs and the optimal positioning of the robots' centers of mass. For example, designs with no leg, two legs, and even more legs are separated into distant clusters. These factors contribute to our model's designs in diverse environments, e.g., placing more weight at the rear end of the robot design to maintain balance while descending stairs and generating robots with more legs to facilitate faster movement on flat planes. 

\begin{figure*}[htbp]
    \centering
    \subfloat[]{%
        \includegraphics[width=0.08\linewidth]{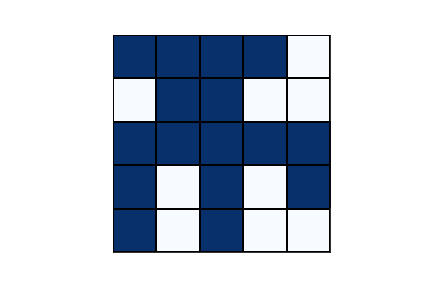}
        \label{fig:ownmodel:subfig1}
    }%
    \subfloat[]{%
        \includegraphics[width=0.08\linewidth]{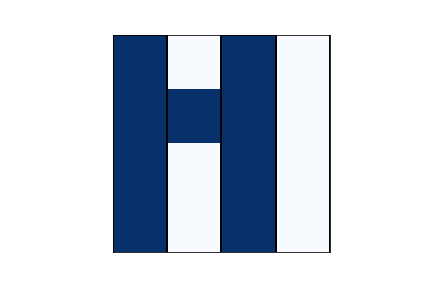}
        \label{fig:ownmodel:subfig3}
    }%
    \subfloat[]{%
        \includegraphics[width=0.08\linewidth]{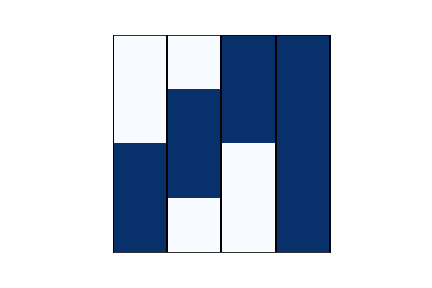}
        \label{fig:ownmodel:subfig5}
    }%
    \subfloat[]{%
        \includegraphics[width=0.08\linewidth]{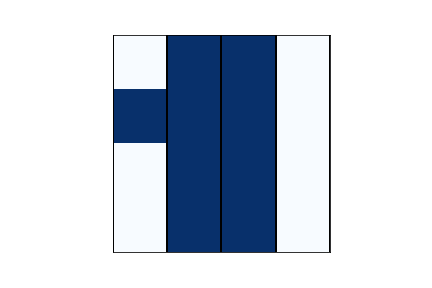}
        \label{fig:compare-related:subfig2}
    }%
    \subfloat[]{%
        \includegraphics[width=0.08\linewidth]{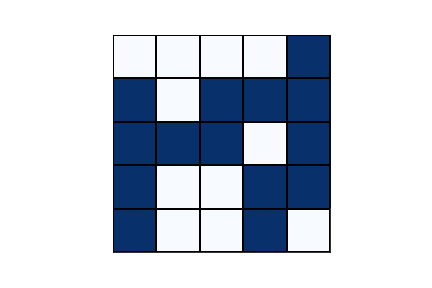}
        \label{fig:compare-related:subfig3}
    }%
    \subfloat[]{%
        \includegraphics[width=0.08\linewidth]{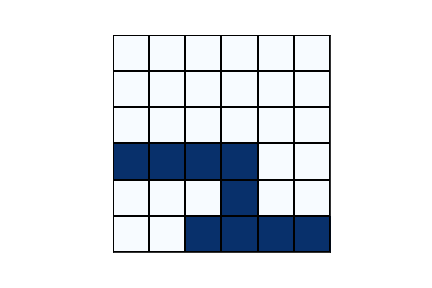}
        \label{fig:downsample:subfig2}
    }%
    \subfloat[]{%
        \includegraphics[width=0.08\linewidth]{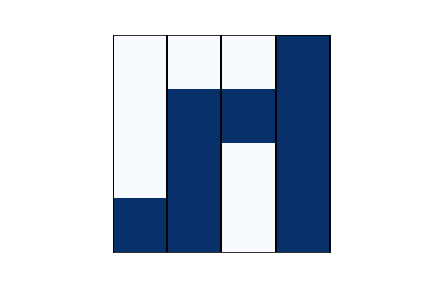}
        \label{fig:downsample:subfig4}
    }%
    \subfloat[]{%
        \includegraphics[width=0.08\linewidth]{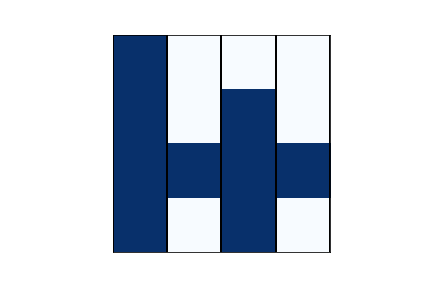}
        \label{fig:downsample:subfig6}
    }%
    \subfloat[]{%
        \includegraphics[width=0.08\linewidth]{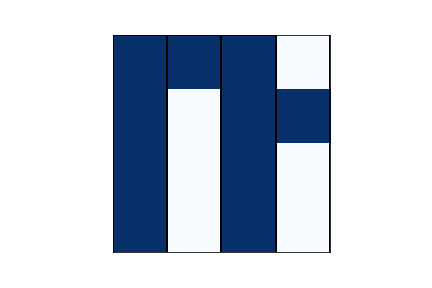}
        \label{fig:12_13:subfig2}
    }%
    \subfloat[]{%
        \includegraphics[width=0.08\linewidth]{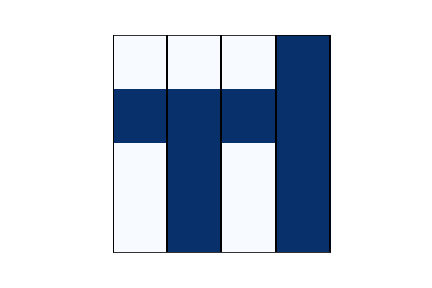}
        \label{fig:12_13:subfig3}
    }%
    \caption{Examples of different models and tasks: (a)-(c) Our model tasks for unidirectional, back-and-forth, and descending locomotion; (d)-(e) No-Compare and No-CLM models; (f)-(h) Downsampled training data examples; (i)-(j) Uni-downstairs and Uni-back tasks.}
    \label{fig:combined_figure_one_line}
    \vspace{-6mm}
\end{figure*}

% [Weicheng&Luyang TODO]: analyze (1) intuitiveness of robot clustering results, e.g., distinctions across clusters, and (2) different clustering standards for different task objectives
We further conduct manual validations to confirm our model's consideration of the task objective and the operating environment in its robot designs, which help it produce highly adaptable robots well-aligned with manual designs. Specifically, for unidirectional locomotion tasks, our model typically generates robots with right-side arms that can adjust the center of mass to accelerate movement (Figure~\ref{fig:ownmodel:subfig1}). In tasks requiring back-and-forth locomotion, the designs are more symmetrical, enabling efficient movement in both directions (Figure~\ref{fig:ownmodel:subfig3}). For the stair-descending task, the center of mass is strategically placed at the rear part of the robots, enhancing balance during the movement (Figure~\ref{fig:ownmodel:subfig5}). 
%Overall, the quality of the generated robots is notable, featuring intuitive and valid designs with functional legs and arms that are akin to manually designed counterparts. Additionally, some models showcase the ability to dynamically increase step size through actuation, enhancing locomotion speed — a feature that has proven inspiring to domain experts, as illustrated in Figure~\ref{fig:ownmodel:subfig1}.

% Further details on the elements contributing to our model's superior performance are discussed in Appendix \ref{sct:discussion}.

% \input{sections/validations}
\section{Discussion} \label{sct:discussion}
\begin{table*}[!ht]
\centering
\tiny
\setlength{\tabcolsep}{3.2pt}
\begin{tabular}{l|ccccc|ccccc|ccccc}
\hline
                         & \multicolumn{5}{c|}{\textbf{Uni-Directional Locomotion}}                                & \multicolumn{5}{c|}{\textbf{Back-and-Forth Locomotion}}                                 & \multicolumn{5}{c}{\textbf{Downstairs Locomotion}}                                       \\ \hline
                         & \textbf{IF}      & \textbf{PS}    & \textbf{OPT}  & \textbf{GEN}      & \textbf{SR}     & \textbf{IF}      & \textbf{PS}    & \textbf{OPT}  & \textbf{GEN}      & \textbf{SR}     & \textbf{IF}      & \textbf{PS}    & \textbf{OPT}  & \textbf{GEN}      & \textbf{SR}     \\ \hline
\textbf{our model}       &\textbf{70.00}\%          & 13.78          & 5.45          & 81.00\%          & \textbf{98.00}\%          & \textbf{61.00\%} & \textbf{11.99} & 2.19          & 74.00\%          & 89.00\%          & \textbf{61.00\%} &  \textbf{16.73}          & \textbf{4.11} & 71.00\%          & \textbf{97.00\%}  \\ \hline
\textbf{no-CLM}          & 41.00\%          & 13.46          & 5.31          & 84.00\%          & 92.00\%          & 33.00\%          & 9.42           & 2.16          & 80.00\%          & 88.00\%          & 36.00\%          & 15.73          & 6.03          & 77.00\%          & 85.00\%           \\ \hline
\textbf{no-Compare}      & 65.00\%          & 13.83          & 5.24          & 83.00\%          & 96.00\%          & 53.75\%          & 8.75           & 2.34          & 73.75\%          & \textbf{93.75\%} & 48.00\%          & 16.15 & 5.46          & 81.00\%          & 93.00\%           \\ \hline
\textbf{uni}             & 69.00\%          & \textbf{14.60} & 5.24          & \textbf{93.00\%} & \textbf{98.00\%} & 53.00\%          & 10.76          & 2.02          & \textbf{85.00\%} & 89.00\%          & 48.00\%          & 12.14          & 5.20          & \textbf{87.00\%} & 76.00\%           \\ \hline
\textbf{back}            & 61.00\%          & 13.41          & 4.82          & 81.00\%          & 91.00\%          & 45.00\%          & 10.21          & 2.07          & \textbf{85.00\%} & 91.00\%          & 48.00\%          & 15.48          & 5.62          & 86.00\%          & \textbf{97.00\%}  \\ \hline
\textbf{downstairs}      & 57.00\%          & 11.82          & 4.64          & 78.00\%          & 90.00\%          & 54.00\%          & 10.48          & \textbf{1.96} & 77.00\%          & 89.00\%          & 44.00\%          & 13.86          & 5.18          & 70.00\%          & 84.00\%           \\ \hline
\textbf{uni+back}        & 69.00\%          & 4.51           & 5.07          & \textbf{90.00\%} & 95.00\%          & 55.00\%          & 1.99           & 2.54          & \textbf{95.00\%} & \textbf{95.00\%} & 48.75\%          & 13.12          & \textbf{5.17} & \textbf{90.00\%} & 92.50\%           \\ \hline
\textbf{uni+downstairs}  & \textbf{73.00\%} & 13.73          & \textbf{4.39} & 89.00\%          & 97.00\% & \textbf{57.50\%} & 4.54           & 2.16          & 67.50\%          & 90.00\%          & 45.00\%          & 8.86           & \textbf{4.40} & 81.67\%          & \textbf{100.00\%} \\ \hline
\textbf{back+downstairs} & 68.00\%          & 9.73           & \textbf{4.47} & 70.00\%          & 88.75\%          & 53.00\%          & 1.83           & 2.41          & \textbf{90.00\%} & \textbf{95.00\%} & 42.00\%          & 15.74          & 5.47          & 86.00\%          & \textbf{97.00\%}  \\ \hline
\textbf{GPT-2}           & \textbf{70.00\%} & \textbf{14.41} & 5.47          & 75.00\%          & \textbf{98.00\%} & 50.00\%          & 10.17          & \textbf{1.90} & 77.00\%          & 90.00\%          & \textbf{53.00\%} & \textbf{17.24} & 6.55          & 76.00\%          & 96.00\%           \\ \hline
\textbf{BART}            & NA               & NA             & NA            & NA               & NA               & NA               & NA             & NA            & NA               & NA               & NA               & NA             & NA            & NA               & NA                \\ \hline
\textbf{0.8T}            & 68.00\%          & 13.68          & 5.00          & 78.00\%          & 94.00\%          & \textbf{62.00\%} & \textbf{10.93} & 2.23          & 73.00\%          & 88.00\%          & 40.00\%          & 14.17          & 6.28          & 75.00\%          & 90.00\%           \\ \hline
\textbf{0.6T}            & \textbf{70.00\%} & 14.02          & 4.76          & 75.00\%          & 96.00\%          & 56.00\%          & \textbf{12.08} & 2.02          & 71.00\%          & 93.00\%          & 48.00\%          & 15.81          & 6.40          & 72.00\%          & \textbf{97.00\%}  \\ \hline
\textbf{0.4T}            & 64.00\%          & \textbf{14.25} & \textbf{4.62} & \textbf{90.00\%} & 97.00\% & 49.00\%          & 10.59          & \textbf{1.75} & 83.00\%          & 89.00\%          & \textbf{58.00\%} & \textbf{17.89} & 5.46          & \textbf{91.00\%} & \textbf{97.00\%}  \\ \hline
\end{tabular}
\caption{Evaluation results in main and ablation experiments. Top 3 scores of each metric are bolded.}
\label{tbl:discussion:result}
% \vspace{-6mm}
\end{table*}

Section \ref{sct:experiments} has proven the overall strength of our proposed approach, and this section delves deeper into the contributing factors of our model's superior performance via a series of ablation studies.
% The ablation studies are designed to verify the necessity of the two training objectives (Section \ref{sct:discussion:objective-ablation}) and the impacts of task diversity in the training process (Section \ref{sct:discussion:task-diversity}), amounts of training data (Section \ref{sct:discussion:data-amount}), and the choice of backbone model (Section \ref{sct:discussion:model-selection}).
% Quantitative evaluation results of all the discussion experiments are assorted in Table \ref{tbl:discussion:result}.

\subsection{Training Objective Ablation Experiments} \label{sct:discussion:objective-ablation}
% Our initial analysis underscores the importance of integrating both the CLM and binary decision-making objectives in training our model. 
As shown in Table \ref{tbl:discussion:result}, ablating the CLM objective (\textbf{no-CLM}) leads to higher failure rates according to the IF and SR metrics and a comparable preference over task optimality, reflected in the PS and OPT scores. This suggests that the ablated model focuses less on representing legal and walkable robot structures but instead prioritizes characteristics beneficial for accelerating the robots' movement.
Conversely, the \textbf{no-Compare} model generally succeeds in designing robots as instructed in the prompts, though it occasionally overlooks the optimality of the designs. 

This trend is supported by our qualitative assessments illustrated in Figure~\ref{fig:compare-related:subfig2}, where the \textbf{no-Compare} model tends to produce more stable and symmetric designs that are, however, challenging to mobilize. On the other hand, the \textbf{no-CLM} model’s creations often demonstrate more effective locomotion despite the high failure rate (Figure~\ref{fig:compare-related:subfig3}).
As such, we claim that incorporating both training objectives is crucial to ensure that the model adheres to instructions while also taking care of design quality, 

% Our first examination pinpoints the necessity of both the CLM and binary decision-making objectives when training robot designing models.
% As shown in Table \ref{tbl:discussion:result}, the \textbf{no-CLM} configuration features much higher failure rates (revealed by the IF and SR measures) and a comparable preference for high-quality robot designs (PS and OPT scores), which is intuitive since the model is trained less for representing robot structures but intensely for favoring higher-quality designs.
% The \textbf{no-Compare} model is, on the contrary, capable of designing robots as instructed in most cases while in some scenarios not caring about the optimality of the robots, as suggested by the lower PS, higher OPT, and relatively high standard deviations of the two scores.
% The same trend is observed in our qualitative examinations shown in Figure \ref{fig:qualitative:objective-ablation}, where the \textbf{no-Compare} model's generations are usually more stable and symmetric while difficult to locomote, while the successful generations of the \textbf{no-CLM} model locomote more efficiently.
% To balance instruction-following with quality assurance, we claim that both training objectives are critical for training our model.

\subsection{Task Ablation Experiments} \label{sct:discussion:task-diversity}
We further investigate the importance of task diversity to the generalizability of our model.
% examine whether training our model using data synthesized with multiple task objectives enhances its generalizability across tasks. 
Specifically, models trained on single tasks (\textbf{uni}, \textbf{back}, and \textbf{downstairs}) often perform well on their specific training objectives while suffer on tasks outside their training data. Introducing a second training objective (\textbf{uni-back}, \textbf{uni-downstairs}, and \textbf{back-downstairs} models) effectively narrows these performance gaps. These observations suggest that incorporating additional task objectives could enhance our models' capacity for solving other untested tasks, helping broaden the scope of robot designs using our model and leading to innovative and effective solutions difficult for experts to design.

Our manual examinations also reveal that the two-task models display better adaptability to the third task which they are not familiar with. For instance, the \textbf{uni-downstairs} model is prone to generating symmetric configuration to facilitate forward and backward walking (Figure~\ref{fig:12_13:subfig2}) and the \textbf{uni-back} model tends to increase weight distribution towards the back for enhanced stability when moving downstairs (Figure~\ref{fig:12_13:subfig3}). These observations are consistent with our quantitative findings, suggesting that training the models with multiple objectives fosters greater generalizability and encourages the models to generate robust and versatile robot designs for complex tasks.

\subsection{Impacts of Training Data Sizes} \label{sct:discussion:data-amount}
% While the incorporation of larger training data could potentially enhance model performance, synthesizing the data is both time- and resource-consuming. 
We conduct repeated training and evaluations of our model using three different subsets of the training data, ranging from 40\% to 80\% in increments of 20\%. The models corresponding to these data subsets are denoted as \textbf{0.4T}, \textbf{0.6T}, and \textbf{0.8T}. Our observations indicate that using less training data results in greater variability in the model’s performance across different environments and task objectives. For instance, the \textbf{0.4T} model outperforms the \textbf{0.6T} and \textbf{0.8T} models in the stair descending task but significantly underperforms both models in the back-and-forth locomotion task. Additionally, manual analysis of the robot designs generated by these models reveals that robot designs produced by the \textbf{0.4T} model are often unstable or impractical for real-world applications, as illustrated in Fig.~\ref{fig:downsample:subfig2}. In contrast, the additional training data used in the \textbf{0.6T} and \textbf{0.8T} models contribute to training models with more stable performance across tasks and settings. It is noteworthy from these results that the performance gain of the \textbf{0.8T} model over the \textbf{0.6T} model is only marginal, suggesting the data efficiency of our proposed approach.

% While more data could potentially benefit training strong models, generating synthesized training data could be time- and resource-consuming. 
% We repeat the training and evaluations of our model using four different sub-portions of the training data from 40\% to 80\%, with a step size of 20\%, the resulting models of which are denoted as \textbf{0.4T}, \textbf{0.6T}, and \textbf{0.8T}.
% % [Weicheng TODO]: finish text after results are out
% We note that the less training data the model uses, the more imbalanced the model's performance is given different environments and task objectives.
% For example, the \textbf{0.4T} model performs drastically differently on the three task objectives, beating the \textbf{0.6T} and \textbf{0.8T} models for the stair descending task objective while lagging behind both other models by large margins for the back-and-forth locomotion objective.
% When analyzing the generated robot designs by the three models, however, we find that the robots generated by the \textbf{0.4T} model are usually unstable or impractical in real robot experiments (illustrated in Figure \ref{fig:qualitative:data-ablation:0.4T}), while the additional training data for the other two models help them generate more pro-task designs.
% Figure \ref{fig:qualitative:data-ablation} provides more example robots generated by the training-data-ablated models.

\subsection{Impacts of Base Model Selection} \label{sct:discussion:model-selection}
Selecting robust backbone models that can comprehend our instructions and generate suitable robot designs is essential in robotics research. Our preliminary experiments utilized two smaller models: \textbf{BART} \cite{bart-orig} and \textbf{GPT-2} \cite{gpt2-orig}, where \textbf{BART} is a masked language model and \textbf{GPT-2} is a CLM. We note that \textbf{BART} struggled across all trials, never producing valid natural language expressions either before or after training. Conversely, the \textbf{GPT-2} model performed well in generating robots for the simpler unidirectional locomotion task but fell short on the other two objectives, highlighting the limitations of smaller models in mastering the complex robot design generation task.
Consequently, we shifted to using a larger LLM, i.e., \textbf{GPT-NEO}, which demonstrated a robust understanding of our prompts and effectively generated accurate, high-quality robot designs. While the exploration of larger LLMs appears promising, in future work it is also important to weigh the substantial demands of training time, evaluation overhead, and data consumption against the performance improvements. 

\section{Conclusion}
We introduced a new approach combining pre-trained language models (LLMs) and differentiable physics simulation tools to automate the design of soft modular robots tailored to specific environments and task objectives. To evaluate the approach, we propose five metrics that assess the quality and generalizability of these designs in different physical contexts. We demonstrate the efficacy of our approach by creating appropriate robots following natural language instructions to accommodate different design tasks.

\section{Limitations and Future work}
One limitation of our current system is the slow cooling time of the Shape Memory Alloy (SMA) coils, which restricts the locomotion speed of the robot. Although SMAs were chosen for their favorable force-to-mass ratio and high work density—making them effective for quickly and efficiently demonstrating our concept—this slow cooling time is a significant drawback for practical applications. To address this limitation, we are exploring the replacement of SMAs with alternative actuators, such as motors, which could increase actuation speed and allow for untethered operation. This change would enhance the adaptability of the modules to more complex environments, making the system more suitable for practical applications.

We used computer vision with color-coded markers to track the $(x, y)$ positions of all robot vertices. However, discrepancies exist between the real and simulated robots: real-world connections have four nodes on the connecting line, while the simulation, limited by DiffTaichi, has only two. Adjusting the simulation would require major changes to DiffTaichi’s core. Perfect alignment remains challenging, especially as simulations show more bounce than real tests, which cannot be easily corrected through parameter tuning.

To address these challenges, we are developing a specialized physical engine tailored to our robots' specific physical properties to enhance simulation fidelity. While the primary objective of this paper is to validate the use of learning-based methods for designing soft modular robots, narrowing the gap between simulated and real-world performance remains a key focus for our future work.

As this is the first work to use natural language for designing soft modular robots, there is no baseline for direct comparison. The closest research involves rigid modular robots, which have smaller search spaces due to fixed block functionality and limited connection options. Thus, we compare our full model to ablated versions—where specific data or objectives are removed—to demonstrate the importance of each component and highlight the model's capabilities.

In future work, we plan to extend our approach to the design of other types of robots, where comparisons with state-of-the-art models in those domains will allow us to further demonstrate the generalizability of our proposed method. However, such comparisons are beyond the scope of this paper, which primarily serves as a proof of concept. Our focus here is to validate the feasibility of using natural language for representing soft modular robot configurations and to explore the potential of pre-trained language models (LLMs) in the robot design process.

\printbibliography
\end{document}